\PassOptionsToPackage{dvipsnames,table}{xcolor}
\documentclass[10pt, a4paper]{dukenlp} %
\usepackage{natbib}

\usepackage[section]{placeins}
\usepackage{subcaption}
\usepackage{graphicx}
\usepackage{booktabs}
\usepackage{listings}
\usepackage{float}
\usepackage{wrapfig}
\usepackage{afterpage}
\usepackage{caption}
\usepackage{dblfloatfix}
\usepackage{needspace}
\usepackage{fnpos}\makeFNbelow %
\usepackage[most]{tcolorbox}
\definecolor{darkblue}{RGB}{0,51,102}

\newcommand{\alexisx}[1]{}

\newcommand{\InPrompt}{No-Log}
\newcommand{\inprompt}{no-log}
\newcommand{\BestScore}{97.4}  %

\tcbset{
  promptbox/.style={
    enhanced,
    colback=gray!3,
    colframe=darkblue,
    boxrule=0.5pt,
    left=6pt, right=6pt, top=6pt, bottom=6pt,
    fontupper=\ttfamily\footnotesize,
    before upper=\setlength{\parskip}{3pt}\setlength{\parindent}{0pt},
    sharp corners,
    breakable,
    before skip=6pt, after skip=6pt,
  }
}

\newtcolorbox{promptbox}[1][]{%
  promptbox,
  title={#1},
  coltitle=white,
  fonttitle=\sffamily\small\bfseries,
  attach boxed title to top left={yshift=-2mm,xshift=4mm},
  boxed title style={colback=darkblue,boxrule=0.5pt,sharp corners},
}

\title{PRO-LONG: Programmatic Memory Enables Long-Horizon Reasoning}

\author{
Alexis Fox$^{1}$, Junlin Wang$^{1}$, Paul Rosu$^{1}$, Bhuwan Dhingra$^{1}$ \\ %
$^{1}$Duke University\\
\texttt{alexis.fox@duke.edu, bdhingra@cs.duke.edu}
}

\begin{document}

\maketitle
\begin{abstract} 
Long-horizon tasks require sustained perception, reasoning, and exploration, and are a persistent challenge for large language model (LLM) agents. This gap is reflected in their limited performance on continual learning benchmarks such as ARC-AGI-3 and NetHack, especially when evaluated without specialized harnesses. Various agent harnesses have been proposed to close the gap, and each commits to a strategy for handling long sequences of observations, i.e., what information to save from the environment and how to load it into model context, a choice we argue is particularly consequential. Existing methods for context management have typically been designed around a tradeoff, where preserving more information makes relevant details harder to retrieve. Recent progress in coding agents, however, makes new approaches to memory viable. We~propose \textbf{PRO-LONG},~a minimal context management framework built around \textit{programmatic memory} for LLM agents in long-horizon, exploratory settings. PRO-LONG addresses the tradeoff by keeping a complete, structured interaction log and using programmatic tools to search this history efficiently. On NetHack, we find increasing gains from using PRO-LONG for cross-episodic memory, with scores reaching up to $\mathbf{2.2}\times$ those of a base coding agent. On the full ARC-AGI-3 public game set, PRO-LONG improves over this baseline agent by an average of $\mathbf{18.0}$ percentage points across frontier models, and matches or exceeds state-of-the-art specialized harnesses (up to $\mathbf{97.4}\%$ best@2) while using $\mathbf{4.2}$--$\mathbf{5.8}\times$ fewer tokens. Relevant code and logs are available at \url{https://github.com/alexisfox7/PRO-LONG}.
\end{abstract}

\section{Introduction}
\label{sec:intro}

Large language model (LLM) agents are being increasingly deployed on more challenging, longer-horizon tasks.
Continual learning benchmarks such as ARC-AGI-3~\citep{arcagi3} and NetHack~\citep{kuttler2020nethack} have subsequently been proposed to evaluate agentic progress, testing whether these agents can infer unknown environment dynamics through exploration and continuously improve over thousands of actions. Out-of-the-box frontier LLMs, evaluated without harnesses or tool access, perform poorly on these settings, achieving limited progression on both ARC-AGI-3 and NetHack~\citep{arcagi3, balrog2025}.

To close this gap, a range of agent harnesses has been proposed for exploratory tasks, wrapping a base model with memory, tools, and other scaffolding, which have shown considerable promise. In parallel, and of particular relevance, agents with access to programmatic tools, such as Claude Code and Codex, have advanced rapidly. Programmatic access enables forms of memory and reasoning that were previously impractical, such as structured search over very large logs. Recent work externalizes long-context processing in a similar spirit~\citep{cao2026coding}, for example through recursive language models that query slices of context through a REPL~\citep{zhang2025rlm}. These trends point toward a broader role for coding agents as context processing systems. \alexisx{1: Revert to earlier version, this sounds ai gen.}

For these agents to maintain long-horizon competence and coherence, how to properly engineer context becomes a central concern. A natural way to frame agent context management is to distinguish between the \textit{accessed} information, through the active context window of the Transformer network, and the \textit{accessible} information, namely memory (i.e., retrievable states such as a log file, saved notes, or an embedding store).
Although modern LLMs can process context windows of a million tokens or more, performance remains expensive and vulnerable to degradation as trajectories grow, a phenomenon often referred to as context rot~\citep{contextrot}. Agents therefore require mechanisms that decide what history remains accessible and how relevant pieces are moved into context.

Proposed agent memory systems can be understood as various implementations of \textsc{read} and \textsc{write} operations over externally stored information~\citep{zhang2024surveymemory}. \textsc{write} is defined by heuristics for what information should be accessible or persisted, whether agent-authored, human-designed, learned, or task-specific. Examples include scratchpad notes, summarized observations, embedded memory entries, and distilled skills or strategies~\citep{park2023generativeagents, packer2024memgptllmsoperatingsystems, chhikara2025mem0buildingproductionreadyai, ouyang2026reasoningbankscalingagentselfevolving, xia2026skillrlevolvingagentsrecursive}. \textsc{read} then determines what is ultimately accessed in context through analogous operations, such as embedding-based retrieval, loading selected memory chunks, keyword search over stored files, or restoring earlier context snapshots~\citep{lewis2020rag, wu2025humanmemory}. This creates a fidelity--tractability tradeoff.
Reducing what is saved at \textsc{write} time makes \textsc{read} easier, but can discard details whose relevance is evident only in hindsight. Over long horizons, the cost of these decisions accumulates, as a lossy summary can matter many turns later. A timeliness question is also raised as to whether these systems are fully capitalizing on, or are maximally compatible with, modern coding agent capabilities.

\begin{figure*}[t]
\centering
\includegraphics[width=\textwidth]{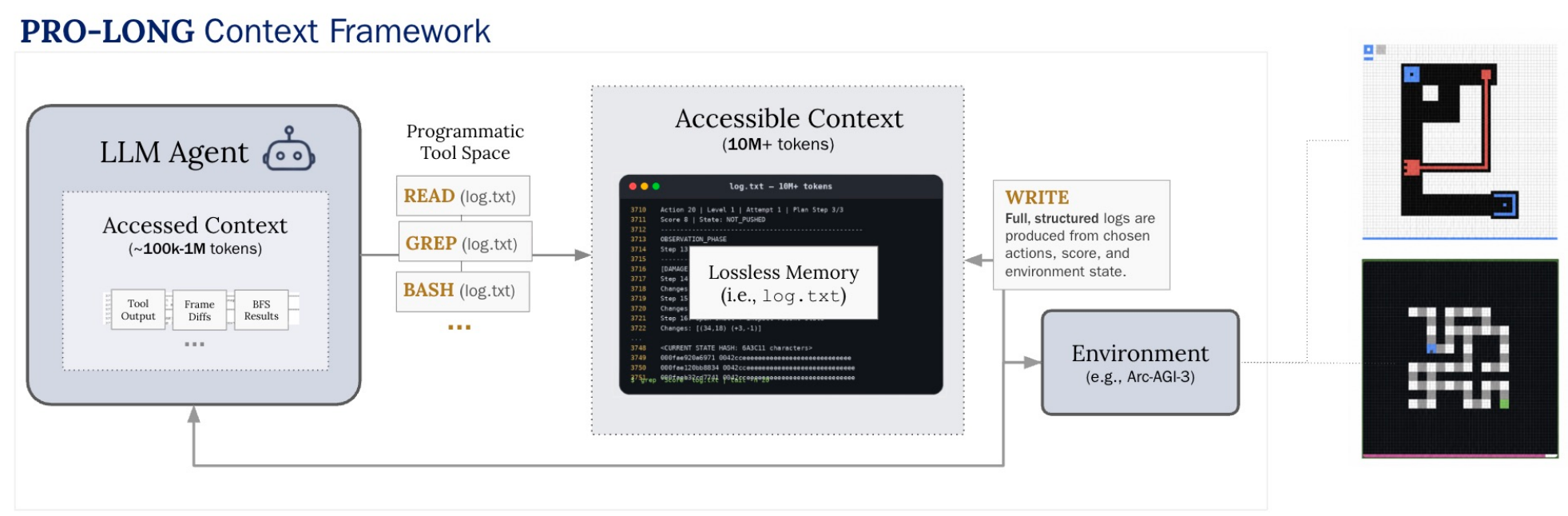}
\caption{Overview of programmatic memory (PRO-LONG) integrated into a basic coding agent setup. The agent has access to a tool space of increasingly programmatic capabilities (\texttt{read}, \texttt{grep}, \texttt{bash}/\texttt{python}), and full, structured logs of environment state are written to \texttt{log.txt}. Note the accessed context (limited to $\sim$100K--1M tokens) versus the far larger accessible context (10M+ tokens). Two example ARC-AGI-3 games are shown on the right: g50t, which requires controlling a ``ghost clone,'' and tu93, a maze game.} 
\label{fig:overview}
\end{figure*} %
We propose \textbf{PRO-LONG}, a context management framework for eliciting \textbf{pro}grammatic, \textbf{long}-horizon reasoning. PRO-LONG centers on our notion of programmatic memory for agents, which we define the \textsc{write} operation as the harness appending every observation, action, and outcome into a log, and the \textsc{read} as programmatic search over that log, primarily through code. We emphasize three motivating design principles. First, \emph{simplicity}; the \textsc{write} needs no learned or heuristic decisions about what information to store, since everything is added as the environment outputs it. \textsc{read} requires no specialized retrieval, e.g., accessing a vector database or index. Second, \emph{losslessness}; nothing is compressed or summarized when writing, so the log is a faithful, ground-truth record of environment state. Third, the approach is \emph{uniquely enabled by, and especially compatible with, coding agents}. Programmatic search, via regular expressions or code, is native to coding agents \citep{cao2026coding}, and this capability keeps retrieval tractable even over very long logs; in practice, we find PRO-LONG retrieval to be tractable for logs over 100K+ lines.

\paragraph{Contributions.} PRO-LONG achieves state-of-the-art results on ARC-AGI-3 and NetHack; we attribute its performance to, and organize our contributions around, the three design principles above. \looseness -1
\begin{itemize}\setlength{\itemsep}{7pt}
\item \textbf{Framework.} We introduce PRO-LONG, a context framework optimized for long-horizon coherence, which introduces programmatic memory for agents as an append-all \textsc{write} to a log and a code-based \textsc{read} over that log. Our design is motivated by simplicity, losslessness of information, and maximal compatibility with coding agent capabilities.
\item \textbf{Benchmark Performance.} Across model families and under matched scoring setups for ARC-AGI-3, PRO-LONG achieves performance comparably to or better than specialized harnesses while using $4.2$--$5.8\times$ fewer billed tokens. With Opus~4.6, it scores $42.4\%$ pass@1, the best public result for that model. With GPT-5.5, PRO-LONG reaches $41.2\%$ pass@1, compared with the prior best of $45.1\%$, while using $5.8\times$ fewer tokens. With Fable~5 at a full action budget, we achieve $\mathbf{\BestScore}\%$ best@$2$ at less than one-third the cost of the strongest prior harness. On NetHack, PRO-LONG's best episode reaches $\mathbf{25.7}\%$ in a single run and $\mathbf{40.8}\%$ within a lifetime, above the best published episode ($20.6\%$). Despite being a minimal addition to standard coding agents, PRO-LONG improves over the base coding agent without a log by $18.0$ points on ARC-AGI-3 and by up to $2.2\times$ on NetHack.

\item \textbf{Drivers of Performance.} We show that adding PRO-LONG to coding agents with even minimal tooling performs strongly, supporting the simplicity principle. The gains are driven by full log access. Performance improves steadily with increasingly powerful programmatic access (Table~\ref{tab:tool-ablation}), while further abstractions, such as persistent workspaces and tools for writing notes, add little (Table~\ref{tab:workspace-ablation}). Improvements are clearest in games where long-horizon reasoning is most useful, i.e., the current board cannot determine the game dynamics (Figure~\ref{fig:pergame_bars}). \looseness -1
\end{itemize}

\section{Setup \& Methods}
\label{sec:setup}

\subsection{Benchmarks and Evaluation}
\label{sec:env}

We evaluate on two challenging continual learning benchmarks, ARC-AGI-3 and NetHack, which stress different aspects of long-horizon reasoning. ARC-AGI-3 has 25 public games of six to ten levels each; the rules are not disclosed, so the agent must infer environment dynamics through exploration. NetHack is a classic dungeon crawler whose conventions an agent is likely to know, but a single game can take over $10{,}000$ actions and ends permanently on death. Therefore, it demands long-horizon planning and, across repeated episodes, the ability to learn from much earlier failures. We detail each environment and its evaluation protocol below.

\paragraph{ARC-AGI-3.} To provide intuition for the games in this benchmark, Figure~\ref{fig:overview} shows example board states from two: \texttt{tu93}, a maze game in which the agent must reach a goal tile while disarming turrets, and \texttt{g50t}, where ``rewinding'' creates a ghost that repeats the agent's previous path, so runs must be planned such that ghosts hold switches open.
For all games, boards are $64\times64$ grids of pixels, each taking one of 16 colors. The action space consists of four directional moves, a click targeting a specific $(x,y)$ cell, undo, and reset. Available actions vary by game, from click-only to the full set, and most games have an action budget or timer that forces a reset when exceeded.

We use the benchmark's standard metric, \emph{Relative Human Action Efficiency} (RHAE)~\citep{arcagi3}. Each level is scored by the agent's action count $a_{l,e}$ against an upper-median human baseline $h_{l,e}$, $S_{l,e} = \min\!\bigl((h_{l,e}/a_{l,e})^{2},\, 1.15\bigr)$, so beating the human baseline earns up to $115\%$. Later levels are weighted more heavily, each environment's score is capped at its weighted fraction of levels completed, and the underlying weighted average over the $|D|$ environments is:
\begin{equation}
T = \frac{1}{|D|}\sum_{e\in D} \frac{\sum_{l=1}^{n} l\,S_{l,e}}{n(n+1)/2}.
\label{eq:score} 
\end{equation}  

\paragraph{NetHack.} We also evaluate on the NetHack Learning Environment (NLE)~\citep{kuttler2020nethack} using the BALROG evaluation harness~\citep{balrog2025}. NetHack is a dungeon exploration game where winning requires descending more than 50 dungeon levels. A successful NetHack run thus often requires tens of thousands of actions. Each run procedurally generates a dungeon, whose map is revealed as the agent explores. The agent must manage food and equipment, as well as fight or avoid monsters; death permanently ends the episode. The BALROG progression score maps the deepest dungeon level and the highest experience level an agent reaches to the empirical probability that a human run at that point goes on to win. The resulting scores range from $0\%$ to $100\%$. In our cross-episodic setting, following prior work on MiniHack (a simplified derivation of NetHack) \citep{nottingham2024sso, ahmed2026theorycoder2} and described further in \S\ref{sec:nethack}, each lifetime consists of five sequential episodes played by the same agent on a fixed seed; PRO-LONG retains its log across episodes. \looseness -1 %

\begin{table}[!h]
\centering\small
\caption{\textbf{Summary of metrics used in our evaluations.} ARC-AGI-3 repeats are independent runs over each game (25 total), while NetHack repeats are sequential episodes, with up to 5 in a lifetime.}
\label{tab:metrics}
\begin{tabular}{lp{5.6cm}p{5.2cm}}
\toprule
 & \textbf{Averaged} & \textbf{Best of} \\
\midrule
\textbf{ARC-AGI-3} & \textit{pass@1:} mean score over independent runs & \textit{best@$k$:} max over $k$ independent runs \\
\textbf{NetHack} & \textit{mean best@$k$:} best score of episodes $1..k$ for each lifetime, averaged over all lifetimes & \textit{max best@$k$:} best of episodes $1..k$, any lifetime \\
\bottomrule
\end{tabular}
\end{table}
\paragraph{Evaluation protocol.} Unless otherwise specified, all evaluated agents run at high model reasoning effort. On ARC-AGI-3, each game allows at most 500 actions, with at most 20 per turn; for all agents that do not use vision, including ours, boards are represented as $64\times64$ text grids with one character per pixel and the 16 colors mapped to ASCII symbols. In NetHack, each episode allows at most 20{,}000 environment actions, submitted in batches of at most 30. The agent receives BALROG's text observation, including the map, status, inventory, and game messages. GPT models use the Codex agent framework and Claude models use Claude Code, each with its standard tool space (e.g., \texttt{read}, \texttt{write}, \texttt{bash}, \texttt{grep}) in a sandboxed container that provides \texttt{python3} with only the standard library. We also evaluate PRO-LONG with Kimi K3 on Mini SWE Agent~\citep{lieret2025minisweagent}, a minimal harness (around 200 lines of code) with a single \texttt{bash} tool.

Evaluating on multiple continual learning benchmarks at once, with multiple frontier models, has little precedent. We see
significant run-to-run variation for most models (\S\ref{sec:general-ablations}), so
Table~\ref{tab:metrics} summarizes the averaged and best-of metrics we report on both benchmarks
whenever compute permits. On NetHack, we evaluate lifetimes, which constitute five dependent runs as sequential episodes. On ARC-AGI-3, repeated runs are independent, with a fresh session and workspace for each game, so no information is transferred across runs or games; we report
$95\%$ bootstrap confidence intervals. Because prior agents were evaluated under different action budgets or scoring versions, we rescore their publicly available runs for a fair comparison.\footnote{For example, Arcgentica's publicly reported
$36.1\%$ was calculated under an earlier ARC scoring version and rises to $39.0\%$ under the
current scorer (v0.97) used for all agents here.}

\subsection{PRO-LONG \& Agent Architectures}
\label{sec:arch}

We build on coding agents, i.e.,\ LLMs operated in a loop with access to a file system and a shell, able to read and write files and execute code as part of their reasoning. Following \S\ref{sec:intro}, we distinguish the agent's \emph{accessed} state, the information in its active context window at a model call (e.g., system prompt, the current board, and recent tool outputs), from its \emph{accessible} state, the external memory it can reach only through tools. Figure~\ref{fig:overview} visualizes how our context framework, PRO-LONG, integrates into a basic coding agent. Below, we define each agent architecture evaluated, including other public baselines. Full prompts  for our agents are provided in Appendix~\ref{app:prompts}.

\paragraph{Base coding agent.} We define a simple coding agent baseline to measure the effectiveness of adding PRO-LONG to a minimal setup. At each turn, the agent reasons over its accessed context of up to $\sim$250K tokens, may call the tools described in \S\ref{sec:env}, and outputs a short sequence of actions. The environment executes these actions and then calls the agent again, with the resulting board added to the user prompt; we ablate this choice in Figure~\ref{fig:nolog_ablation}. Earlier boards may remain in the active context window, but the baseline provides no external memory beyond what the agent writes itself, e.g.,\ notes or saved boards in workspace files; in practice, it relies heavily on such workspace writing (Table~\ref{tab:tool-usage}). We denote this baseline \emph{\inprompt{}}  and use it as our primary point of comparison. This baseline is stronger than the standard agents in continual learning benchmarks, which typically are only given the current state and a short history, without coding tools. For example, ARC-AGI-3 reference agents are given the last 3--5 board states through a 10-message window~\citep{arcagi3}, and BALROG agents receive the last 16 observations~\citep{balrog2025}.

\begin{figure}[t!]
\centering
\includegraphics[width=0.85\textwidth]{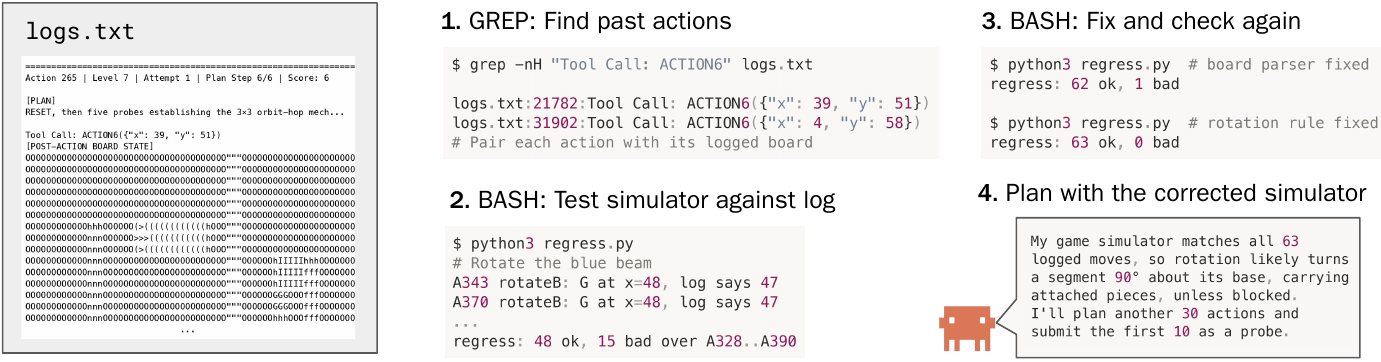}
\caption{\textbf{Example PRO-LONG workflow} (Fable 5, game \texttt{s5i5}, level 7). PRO-LONG writes a Python script (via \texttt{bash}) to replay actions recorded in \texttt{logs.txt} through its coded game simulator, compare the predicted boards with the recorded observations, and revise an inferred game rule. Excerpts are abridged for readability.}
\label{fig:workflow}
\end{figure}
\paragraph{PRO-LONG.} We add PRO-LONG, our minimal context framework, to the base coding agent above. Figure~\ref{fig:overview} shows the corresponding \textsc{write} and \textsc{read} operations for this harness. For each action, the \textsc{write} appends a structured entry \texttt{logs.txt} containing a header (the action number, level, attempt, and score), the agent’s summarized plan, the selected action, and the resulting board. The \textsc{read} searches over this log with the programmatic tools, i.e., regular expressions or code, of effect we discuss in \S\ref{sec:general-ablations}. With these tools, the agent can, for instance, \texttt{grep} the log for score changes to identify the most consequential actions for a level. An interesting example we see is a PRO-LONG agent that writes a \texttt{regress.py} script to replay every action in \texttt{log.txt} and verify its coded game model predicts the logged board states (Figure~\ref{fig:workflow}). We go over more examples on specific games in \S\ref{sec:general-ablations}. We also study an intermediate condition for PRO-LONG that truncates the log to the last 25 actions (\S\ref{sec:benchmark}). We use no subagents, and our prompt is only around 30 lines (Appendix~\ref{app:system-fulllog}).

\paragraph{Baselines.} For ARC-AGI-3, in addition to our base coding agent, \textit{\inprompt{}}, we compare against WorldModeler\footnote{The released agent of \citet{rodionov2026worldmodels} is named \texttt{baseline1}; we refer to it as WorldModeler for clarity, given the executable world model emphasis in their implementation.}~\citep{rodionov2026worldmodels} (GPT-5.5), Arcgentica~\citep{symbolica2026arc} (Opus 4.6), and the Schema harness~\citep{schema2026} (Opus 4.8 and Fable 5). We focus on these three as the strongest public harnesses for the frontier models evaluated, that also follow the standard assumptions of ARC's official leaderboard, including independent runs with no information shared across games. 

WorldModeler and Schema both focus on having the agent create coded ``world models'' of the games. WorldModeler, a harness for Codex, has a prompt of around 600 lines instructing the agent to write and continually update a Python simulator of the game. The harness provides helper scripts and two subagents for tasks like frame preprocessing and animation analysis. WorldModeler is also the only agent evaluated that uses vision. Schema follows a similar approach on Claude Code, replacing the default tools with twelve custom ones. Across turns, the Schema agent updates a \texttt{step(grid, action)} function predicting the next board, calls a backtest tool to validate the model, and searches for plans by breadth-first search. Arcgentica is built on a custom agent SDK, and defines a main orchestrator agent that delegates to four subagents (Explorer, Theorist, Tester, and Solver), which all update a shared memory database of hypotheses and findings.

\needspace{4\baselineskip}%
On NetHack, our references are BALROG's official agent (the leaderboard configuration, given the last 16 observations and no tools) run with the same models as ours, and Glyphbox~\citep{kenforthewin2026nethack}, a NetHack-specific harness exposing a high-level game API for pathfinding and auto-exploration, alongside observation masking, a sliding context window, and explicit note-keeping.

\begin{figure*}[t]
\centering
\includegraphics[width=0.9\textwidth]{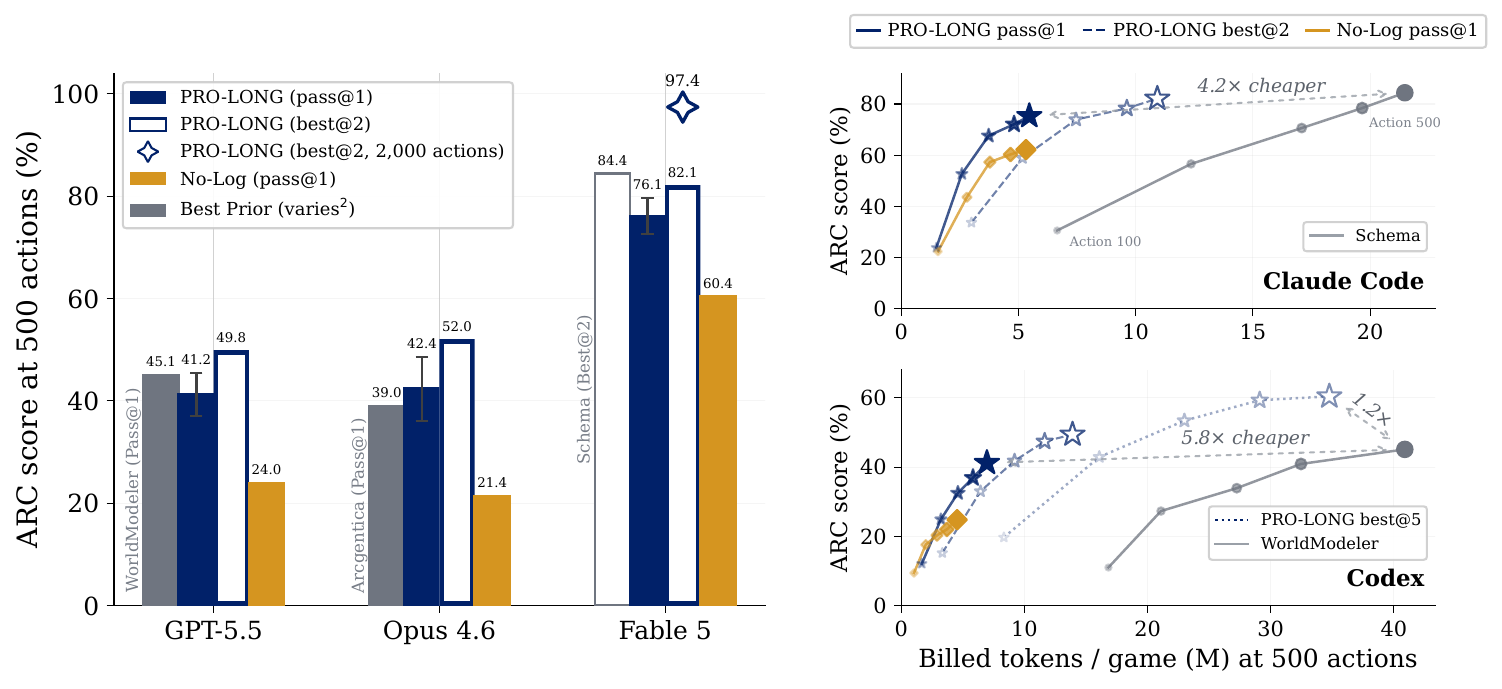} %
\caption{\textbf{PRO-LONG matches or exceeds state-of-the-art ARC-AGI-3 results at $4.2$--$5.8\times$ lower token cost.} \emph{Left:}~ARC-AGI-3 results on the public game set with a 500 action limit, grouped by model and harness. Filled bars show pass@1, with bootstrap confidence intervals where multiple runs are available (five for Codex, two for Claude Code). Outlined bars show best@$k$. We rescore the released runs of WorldModeler, Arcgentica, and Schema for a consistent comparison (see footnote~\ref{fn:selection}). Schema reports best@2; the others report pass@1. \emph{Right:} Performance versus billed tokens per game for PRO-LONG and the strongest prior harness on Codex and Claude Code, from 100 to 500 actions. PRO-LONG stays within $2$--$4$ points of the strongest prior harness at $4.2$--$5.8\times$ lower cost.} %
\label{fig:main_results}
\end{figure*}

\section{Main Results}
\label{sec:ablations}

We evaluate PRO-LONG on the full ARC-AGI-3 public game set and on NetHack. Despite being a minimal addition to a basic coding agent, PRO-LONG improves significantly over the \inprompt{} baseline on both, and on ARC-AGI-3 matches or outperforms specialized harnesses at much lower token cost. We present benchmark performance (\S\ref{sec:benchmark}), study cross-episodic learning on NetHack (\S\ref{sec:nethack}), and ablate which components drive performance and analyze agent behavior qualitatively (\S\ref{sec:general-ablations}).

\subsection{ARC-AGI-3 Benchmark Performance}
\label{sec:benchmark}
Figure~\ref{fig:main_results} compares benchmark performance for PRO-LONG and the relevant baselines discussed in \S\ref{sec:arch}. Across all frontier models: GPT 5.5, Opus 4.6, and Fable 5, PRO-LONG consistently improves by $15.7$--$21.0$ percentage points over the coding agent baseline (\inprompt{}, pass@1). Further intuition for this result, and how PRO-LONG incentivizes long-horizon reasoning, are in \S\ref{sec:general-ablations}. 

Under matched scoring setups,\footnote{\label{fn:selection}For prior agents, the selection procedure is not always public and replicates are not available. WorldModeler evaluates a single playthrough per game, and Arcgentica releases one run per game with no stated selection procedure; we label both pass@1. Schema uses a best-of-two heuristic over Opus 4.8 and Fable 5, which we compare against our best@2.} PRO-LONG either surpasses specialized harnesses (Opus 4.6, +$3$pp at pass@1) or is within comparable range ($2$--$4$pp on GPT-5.5 and Fable 5). We emphasize that PRO-LONG’s simple design is model-agnostic in implementation, whereas prior harnesses are typically tuned to a specific model. This simplicity also translates into substantial token savings.

In Figure~\ref{fig:main_results}, we also compare the billed token cost of our method against the best prior harness on each coding agent, Claude Code and Codex. On Codex, PRO-LONG reaches $41.2$ pass@1 versus WorldModeler's $45.1$ while using $\mathbf{5.8}\times$ fewer tokens. At best@5, it reaches $60.1$ while remaining $\mathbf{1.2}\times$ cheaper. On Claude Code, PRO-LONG reaches $82.1$ best@2, close to Schema's $84.4$ best@2 while using $\mathbf{4.2}\times$ fewer tokens. Across the full game set, this corresponds to savings of roughly $\mathbf{400}$ \textbf{million tokens}.

Beyond the matched $500$-action comparison, we also run Fable 5 at a $2{,}000$-action budget, where PRO-LONG scores $94.6$ pass@1 at a cost of \$1{,}500 ($150$M billed tokens). PRO-LONG reaches $\mathbf{\BestScore}\%$ best@2 at a total cost of \$1{,}750; this is a lower bound on best@2, as several games were not rerun. A few difficult games account for most of the cost; \texttt{bp35} alone costs \$298. Schema releases logs only for its retained best runs (up to 3{,}000 actions),\footnote{\label{fn:retained}Schema releases logs only for each game's retained run, i.e., the better of its up to two attempts, so our cost comparisons for Claude use retained run costs. Even comparing the full cost of our best@2 (both runs) against Schema's single retained runs, PRO-LONG remains $2.0\times$ cheaper. For Codex, we instead report the full best@5 cost (all five runs) to emphasize the savings relative to WorldModeler's pass@1.} whose reported cost totals \$6{,}447 for $99.0$ best@2; even this partial total is more than $\mathbf{3}\times$ higher than ours.

\begin{figure}[t]
\centering
\begin{minipage}[c]{0.40\textwidth}
  \centering
  \includegraphics[width=0.87\linewidth]{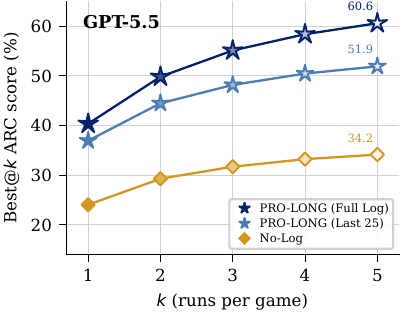}
  \caption{\textbf{Best@$k$ comparison} for PRO-LONG, a truncated version (log keeps only the last 25 actions), and \inprompt{}. The gap widens with $k$.}
  \label{fig:bestk}
\end{minipage}\hfill
\begin{minipage}[c]{0.56\textwidth}
  \centering
  \includegraphics[width=\linewidth]{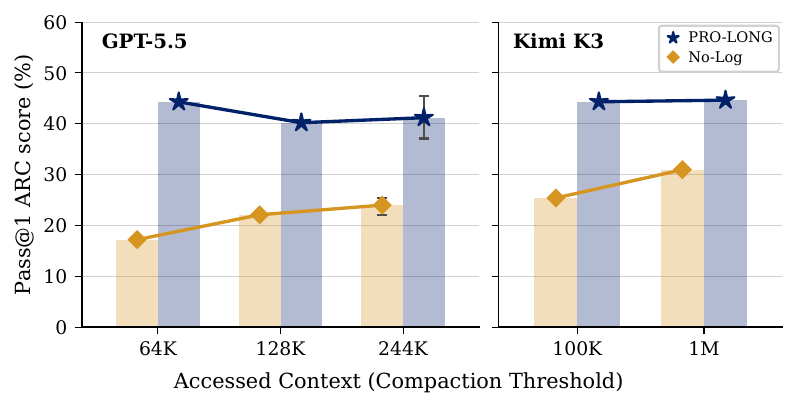}
  \caption{\textbf{Accessed versus Accessible Context.} Decreasing accessed context does not impair PRO-LONG, while no-log degrades. \textit{Left:} GPT 5.5 + Codex. \textit{Right} Kimi + Mini-SWE. ($n=5$ for 244K, $n=1$ otherwise).} %
  \label{fig:context_access}
\end{minipage}
\end{figure}

In Figure~\ref{fig:bestk}, we do further best@$k$ analysis, looking at how score improves as $k$ increases and how variants of PRO-LONG compare. Note, first, there is significant variance in game outcomes across independent runs, and thus it is important to report best@$k$ results where possible.  For example, GPT-5.5 reaches $41.2$ at pass@1, while best@5 rises by $\mathbf{18.9}$ percentage points to $60.1$. Second, increasing the ``fullness'' of memory leads to a large improvement in score, and this gap widens with $k$. The widening gap suggests that fuller memory covers more games, i.e., expands the set of games solvable at all, and that games requiring long-horizon reasoning are harder in general, with higher variance that may
reflect interesting exploration failure modes.

\FloatBarrier

\subsection{Cross-Episodic Learning on NetHack}
\label{sec:nethack}

\begin{figure}[!t]
\centering
\includegraphics[width=0.47\textwidth]{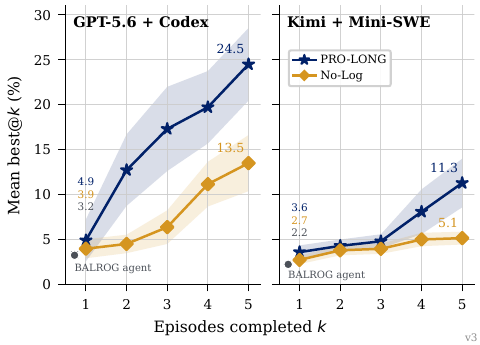}\hfill
\includegraphics[width=0.47\textwidth]{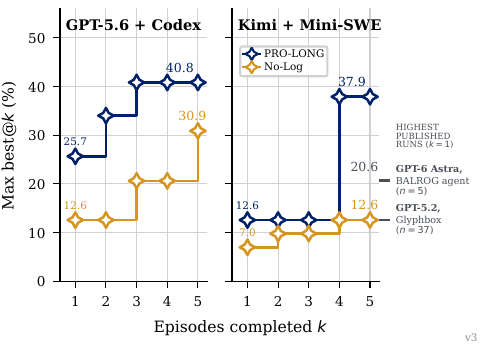}
\caption{\textbf{PRO-LONG shows increasing gains on NetHack with cross-episodic memory.} BALROG progression over the first $k$ episodes of a lifetime for GPT-5.6 + Codex ($n{=}10$) and Kimi + Mini-SWE ($n{=}15$ each). \textit{Left:} Mean of the best score within the first $k$ episodes, $\pm$ SE across lifetimes, with the baseline BALROG agent shown in gray at $k{=}1$. \textit{Right:} Best episode observed within the first $k$ episodes of any run, with the highest published single episode runs marked on the right.\looseness -1} 
\label{fig:nethack-bestk}
\end{figure}

We complement our ARC-AGI-3 results by evaluating on NetHack, as scored by BALROG progression~\citep{balrog2025}. NetHack episodes can be far longer---runs exceed 10,000 environment actions in our tests---and are therefore a stress-test of long-horizon competency. A single episode has significant variance due to the permanent death mechanic, which makes NetHack a suitable setting for testing explicit \textit{cross-episodic memory}. This setting retains information from one episode into later ones, and in particular can test whether a weak early episode is recoverable (which we find it can; Figure~\ref{fig:nethack-lifetimes}). Similar cross-episode tests have been previously studied on simplified MiniHack (NetHack-derived) tasks~\citep{nottingham2024sso, ahmed2026theorycoder2}. 

To test cross-episodic memory,  we define \textit{lifetimes} which consists of five sequential episodes played by the same agent. PRO-LONG persists its log across episodes, such that the accessible history spans a full lifetime, which is empirically over 24{,}000 cumulative actions. At $k=1$, before any cross episode history is available, all agents are weak (including the base BALROG agent), with PRO-LONG only showing a modest improvement over no-log. After five episodes, the average best@$k$ progression reaches $\mathbf{24.5}\%$ versus $13.5\%$ with GPT 5.6 and $\mathbf{11.3}\%$ versus $5.1\%$ with Kimi, approximately twofold higher in both settings (Figure~\ref{fig:nethack-bestk}). We find that PRO-LONG with GPT 5.6 improves increasingly and steadily; with Kimi, we see that performance is uniformly weak on early episodes across all harnesses, while PRO-LONG jumps later in score (max best@$k$ $12.6\%\rightarrow 37.9\%$) in episode four and five.

Released BALROG leaderboard episodes reach at most $20.61\%$ (GPT-6 Astra, best of $n{=}5$ independent episodes), and Glyphbox's best GPT-5.2 episode reaches $12.56\%$ ($n{=}37$)~\citep{kenforthewin2026nethack}. Even without cross-episode memory, PRO-LONG's strongest first episode (GPT-5.6 + Codex) reaches $\mathbf{25.66}\%$, above the prior best of $20.61\%$. With it, the best Codex lifetime reaches $\mathbf{40.8}\%$, and even a Kimi lifetime reaches $37.9\%$, above every public episode.

\FloatBarrier

\subsection{General Ablations} 
\label{sec:general-ablations}

\begin{table}[t]
\begin{minipage}[t]{0.46\textwidth}
\centering
\small
\caption{\textbf{Tool ladder ablation along the hierarchy of programmatic access (GPT 5.5).} Increasing level of programmatic control over logs and programmatic reasoning lead to improved performance, whereas other types of tooling such as write do not as much.} %
\label{tab:tool-ablation}
\begin{tabular}{lc}
\toprule
\textbf{Tools enabled} & \textbf{Score (\%)} \\
\midrule
Read only            & 23.1 \\
$+$ Grep / Regex Search    & 27.2 \\
$+$ Python    & 38.3 \\
$+$ Write, Edit            & 41.2 \\
\bottomrule
\end{tabular}
\end{minipage}\hfill
\begin{minipage}[t]{0.50\textwidth}
\centering
\small
\caption{\textbf{Workspace persistence ablation.} pass@1 (\%) for GPT 5.5 with bootstrap confidence intervals ($n=5$ persistent, $n=2$ cleared). Clearing the workspace after every call has little effect on PRO-LONG but reduces \inprompt{} performance by about four points. }
\label{tab:workspace-ablation}
\begin{tabular}{lcc}
\toprule
\textbf{Workspace} & \textbf{PRO-LONG} & \textbf{\InPrompt{}} \\
\midrule
Persistent & 41.2 {\footnotesize $\pm$3.5} & 24.0 {\footnotesize $\pm$2.0} \\
Cleared every call & 40.7 {\footnotesize $\pm$3.6} & 19.9 {\footnotesize $\pm$2.1} \\
\bottomrule
\end{tabular}
\end{minipage}
\end{table}

\paragraph{PRO-LONG works with limited accessed context.} We observe that \texttt{logs.txt}, PRO-LONG's primary source of accessible context, can grow up to 320K lines long, while any given turn accesses only a small fraction. The accessed context is itself controllable, with direct implications at inference time (Table~\ref{tab:token-cost}). We ablate it by lowering Codex's compaction threshold from 244K to 64K. PRO-LONG does not degrade, scoring slightly higher at 64K than at the default ($44.3\%$ vs.\ $41.2\%$, within its confidence interval), while \inprompt{} falls monotonically. We find the same trend in a simpler setting, using Kimi and the Mini-SWE harness; reducing the threshold from 1M to 100K, PRO-LONG stays flat, within $0.3$ points, while \inprompt{} drops more than five.

\paragraph{Programmatic access drives performance.} PRO-LONG's \textsc{read} operation is defined by a ``programmatic tool space.'' We would expect (1) more capable tools to make the log more useful, and (2) relevant tools that capitalize on the log, rather than other forms of memory, to show the largest performance gains. Table~\ref{tab:tool-ablation} tests, and supports, both these hypotheses. Starting from the full standard tool space ($41.2\%$), removing \texttt{write} and \texttt{edit} tools lowers performance by only $2.9$ points, even though these are the agent's primary means of maintaining other forms of memory, like notes. Programmatic tools account for most of the remaining gain. Starting from file reading alone ($23.1\%$), adding regular expression search such as \texttt{grep} improves performance by $4.1$ points, while adding Python access contributes more than ten additional percentage points.

Note that since PRO-LONG builds on a base coding agent, it can \textsc{write} other forms of memory beyond the log, such as self-authored notes. A core property of how useful such memory is relies on whether it is persistent, i.e.,\ notes must be maintained across model calls so the agent can refer back to them. We ablate how much PRO-LONG is affected by disabling the persistence of these other forms of accessible context, through wiping the workspace every model call. We find that for the \inprompt{} agent, which relies heavily on self-authored notes and helper functions (Table~\ref{tab:tool-usage}), performance degrades by roughly four points, while PRO-LONG is essentially unaffected (Table~\ref{tab:workspace-ablation}). \alexisx{2: Add a small statement of what's the takeaway from this result} %

\begin{table}[b]
\centering
\caption{\textbf{Where programmatic memory helps} (GPT 5.5). \emph{Left:} Pass@1 on the four games discussed in \S\ref{sec:general-ablations}, bars span score range over $n=5$ runs. \emph{Right:} Tool call distribution across the 25 games. PRO-LONG spends most of its calls on programmatic analysis of the log ($40.8\%\to60.6\%$), while \InPrompt{} spends them on workspace management ($19.1\%\to59.2\%$).} %
\label{tab:tool-usage}
\begin{minipage}[c]{0.29\linewidth}
  \centering
  \includegraphics[width=\linewidth]{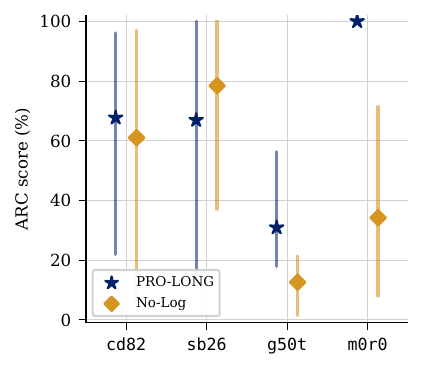}
\end{minipage}\hfill
\begin{minipage}[c]{0.68\linewidth}
  \centering\footnotesize
  \setlength{\tabcolsep}{4pt}
  \begin{tabular}{l@{\hskip 4pt}p{2.6cm}rr}
  \toprule
  \textbf{Category} & \textbf{Tools} & \textbf{PRO-LONG} & \textbf{\InPrompt{}} \\
  \specialrule{\lightrulewidth}{\aboverulesep}{0pt}
  \rowcolor{gray!12} Programmatic analysis & \texttt{python3} & 60.6\% & 40.8\% \\
  \addlinespace
  \rowcolor{gray!12} Log parsing & on \texttt{logs.txt} & 20.3\% & 0.0\% \\
  \quad Recent state & \texttt{tail}, \texttt{head}, \texttt{cat} & 10.7\% & -- \\
  \quad Search across turns & \texttt{grep}, \texttt{sed}, \texttt{awk} & 9.6\% & -- \\
  \addlinespace
  \rowcolor{gray!12} Workspace management &  & 19.1\% & 59.2\% \\
  \quad Navigation & \texttt{find}, \texttt{ls}, \texttt{pwd} & 14.6\% & 29.0\% \\
  \quad Writing & \texttt{apply\_patch}& 4.6\% & 30.3\% \\
  \bottomrule
  \end{tabular}
\end{minipage}
\end{table}

\paragraph{Qualitative Analysis.} We show individual game score statistics over five runs for PRO-LONG and our base coding agent in Figure~\ref{fig:pergame_bars}, which helps explain where programmatic memory provides the greatest benefit. Both agents perform strongly on some games, such as \texttt{sb26} and \texttt{cd82}, where the current board largely specifies the puzzle dynamics, so access to the full trajectory provides little additional information. These were also the shortest consistently completed games: successful runs required median totals of 157 and 213 actions for PRO-LONG, and 174 and 271 actions for \inprompt{}, respectively.

Programmatic memory is most useful for games that require longer-horizon reasoning, like \texttt{g50t} and \texttt{m0r0}. For \texttt{g50t} (Figure~\ref{fig:overview}), individual PRO-LONG run logs exceeded 320{,}000 lines. The primary game mechanic is ``rewinding,'' which creates ghosts that replay the agent's previous paths; later levels require coordinating multiple ghosts as they move and teleport toward switches. Even with a lengthy log and complex game dynamics, PRO-LONG reaches $56.3\%$ with GPT 5.5. For \texttt{m0r0}, the agent controls two blocks in a maze-like game. Although our setup does not require or emphasize ``world models,'' PRO-LONG codes one on its own, defining transition functions that grow with later levels (e.g., by adding a switch state variable) and running breadth-first search over joint block positions to find valid routes of more than $50$ actions. PRO-LONG scores $100\%$ on all five runs, compared with $34.2\%$ for the \inprompt{} baseline. Tool use across the full set of games (Table~\ref{tab:tool-usage}) supports a similar attribution, as PRO-LONG spends most of its calls on programmatic analysis of the log, while \inprompt{} spends them on writing files.

\FloatBarrier
\section{Related Work} 

\paragraph{Context Management.}
Several studies have pointed out that performance
degrades as context length increases,
a phenomenon labeled as ``context rot''
\citep{liu2024lost,contextrot}.
One line of work addresses these limitations through externalized memory systems that store information outside the active context and retrieve it when needed \citep{mei2025surveycontextengineeringlarge}. These approaches differ by how they write, read and synthesize information. MemGPT \citep{packer2024memgptllmsoperatingsystems} introduces a tiered memory system that moves information in and out of a fixed context window. Similarly,  Mem0 \citep{chhikara2025mem0buildingproductionreadyai} dynamically extracts, consolidates, and retrieves information from current context. ReasoningBank \citep{ouyang2026reasoningbankscalingagentselfevolving} turns past experience into entries,
whereas SkillRL \citep{xia2026skillrlevolvingagentsrecursive} converts trajectories into reusable skills.

\paragraph{Programmatic Harnesses and ARC-AGI-3} A new line of work on context management has emerged that stores all information and accesses it dynamically. For example, \cite{cao2026coding} uses coding agents for long-context processing. Moreover, recursive language models \citep{zhang2025rlm} use code to query parts of a long input through a REPL. Specifically, for long-horizon tasks like ARC-AGI-3, RLM-inspired harnesses represent game states programmatically to form and revise hypotheses before selecting actions \citep{symbolica2026arc, karten2026prime, nvidia_oo_agents_2026}. World-model methods build an executable Python model of each game and plan with it before acting \citep{rodionov2026worldmodels, schema2026, courtis2026opineworldprogrammaticworldmodeling, sarafian2026workspaceoptimizationtrainagent}. PRO-LONG externalizes and tracks the agent’s own trajectory: it maintains a complete, append-only log that the agent queries with code as needed.

\section{Conclusion}
\label{sec:conclusion}
We introduce \textbf{pro}grammatic memory for \textbf{long}-horizon reasoning with our framework PRO-LONG. We claim that (1) programmatic perception and reasoning, such as using code to parse board grids, model game transitions, and plan with breadth-first search, are broadly effective for continual learning tasks, and (2) simple harnesses can elicit these behaviors, with PRO-LONG a minimal but highly effective instance. We emphasize simplicity, alongside lossless memory and compatibility with coding agents. Even with a minimal prompt and agent architecture, PRO-LONG frequently creates sophisticated models of game dynamics and long-horizon action plans. As such, our model-agnostic method is competitive across all evaluation settings: on ARC-AGI-3, it achieves up to \BestScore\% on the public benchmark while using significantly fewer tokens than the strongest prior harness on each model; on NetHack, it reaches $25.7\%$ in a single episode, above  the best prior published run.

We evaluate across various frontier models, two of which were released after ARC-AGI-3. Although this may advantage agents using those models, including the baselines, ARC-AGI-3 remains useful for relative comparison of long-horizon reasoning and context management methods. We also note the significant variance we see across game runs for both benchmarks, and thus report best@$k$  where possible. Given the high cost of evaluating on continual learning benchmarks, an important direction is to see if reinforcement learning and improved harnesses can close the gap between pass@1 and best@$k$. Other exciting directions include co-training models with the harness, and studying how programmatic memory interacts with other memory mechanisms.

\section{Acknowledgments}

We gratefully acknowledge the ARC Prize Foundation for providing computational resources and credits used in this work, as well as for valuable discussions and feedback. Junlin Wang is partially supported by through Grant G-23- 2137070 from the Learning Engineering Virtual Institute to the University of Florida and its partner institutions, including Duke.

\bibliographystyle{plainnat}
\bibliography{references}

\clearpage
\appendix

\section{Scoring and Token Cost Details}
\label{app:cost}

Games are run and scored with the official \texttt{arc\_agi} engine, with scores computed according to Equation~\ref{eq:score} and game versions pinned to those listed in Table~\ref{tab:game-versions}.

\begin{table}[h]
\centering
\footnotesize
\caption{Game versions used for all runs in our experiments.}
\label{tab:game-versions}
\setlength{\tabcolsep}{4pt}
\begin{tabular}{ccccc}
\toprule
\texttt{ar25-0c556536} & \texttt{bp35-0a0ad940} & \texttt{cd82-fb555c5d} & \texttt{cn04-2fe56bfb} & \texttt{dc22-fdcac232} \\
\texttt{ft09-0d8bbf25} & \texttt{g50t-5849a774} & \texttt{ka59-38d34dbb} & \texttt{lf52-271a04aa} & \texttt{lp85-305b61c3} \\
\texttt{ls20-9607627b} & \texttt{m0r0-492f87ba} & \texttt{r11l-495a7899} & \texttt{re86-8af5384d} & \texttt{s5i5-18d95033} \\
\texttt{sb26-7fbdac44} & \texttt{sc25-635fd71a} & \texttt{sk48-d8078629} & \texttt{sp80-589a99af} & \texttt{su15-1944f8ab} \\
\texttt{tn36-ef4dde99} & \texttt{tr87-cd924810} & \texttt{tu93-0768757b} & \texttt{vc33-5430563c} & \texttt{wa30-ee6fef47} \\
\bottomrule
\end{tabular}
\end{table}

We report cost in billed tokens. Each token class is weighted by its list price relative to one uncached input token of a reference model. Multiplying a run's billed token count by the reference input price therefore recovers its dollar cost. Token counts are taken from the usage reported by each API call and summed over the run.

\textbf{Claude models} (reference: Fable~5 input, \$10/M):
\begin{equation}
  \mathrm{billed}
  =
  \mathrm{input}
  + 5\,\mathrm{output}
  + 0.1\,\mathrm{cache\_read}
  + 2\,\mathrm{cache\_write},
  \label{eq:billed-claude}
\end{equation}
where \texttt{input\_tokens} already excludes cached tokens. Cache writes use the one-hour rate because Claude Code issues only one-hour-TTL writes; we verified that \texttt{ephemeral\_5m}${}=0$ across our runs. Opus has the same relative prices, so the same weights apply. 

\textbf{Codex models} (reference: GPT-5.5 input, \$5/M):
\begin{equation}
  \mathrm{billed}
  =
  (\mathrm{input}-\mathrm{cached})
  + 6\,\mathrm{output}
  + 0.1\,\mathrm{cached},
  \label{eq:billed-codex}
\end{equation}
No cache-write term is included because GPT-5.5 does not charge separately for cache writes.

All token comparisons in the paper use the same model family, so relative billed token counts do not depend on the chosen reference model.

\section{Additional Ablations and Analysis}
\label{app:additional}

\paragraph{Decreasing accessed context has varying effects on token cost.} To complement our compaction analysis in Figure~\ref{fig:context_access}, we report what ablating the accessed context costs in terms of billed tokens. Table~\ref{tab:token-cost} reports token usage at each compaction threshold, with $Billed$ computed by Equation~\ref{eq:billed-codex}. Lowering this threshold nearly halves total input, from $46.4$M to $24.3$M tokens per game, but leaves billed cost roughly unchanged, as the savings are offset by a lower cache hit rate. Cached input decreases from $97.3\%$ of $Input$ at the default to $90.7\%$ at 64K.

\begin{table}[t]
\centering
\footnotesize
\setlength{\tabcolsep}{7pt}
\caption{\textbf{Token cost across compaction thresholds} (GPT 5.5, PRO-LONG), in millions of tokens per game. Default setting is averaged over $n=5$. The billed token total adds the uncached input $Input-Cached$ to the weighted output $6Ouput$ and cached input $0.1Cached$. }
\label{tab:token-cost}
\begin{tabular}{lrrrrr}
\toprule
\textbf{Context setting} & $Input$ & $Cached$ & $Input-Cached$ & $Output$ & $Billed$ \\
\midrule
64K             & 24.3 & 22.0 & 2.3 & 0.4 & \textbf{6.9} \\
128K            & 30.5 & 29.1 & 1.3 & 0.3 & \textbf{6.0} \\
Default (244K)  & 46.4 & 45.1 & 1.3 & 0.2 & \textbf{7.1} \\
\bottomrule
\end{tabular}
\end{table}

\begin{figure}[t]
\centering
\includegraphics[width=0.8\linewidth]{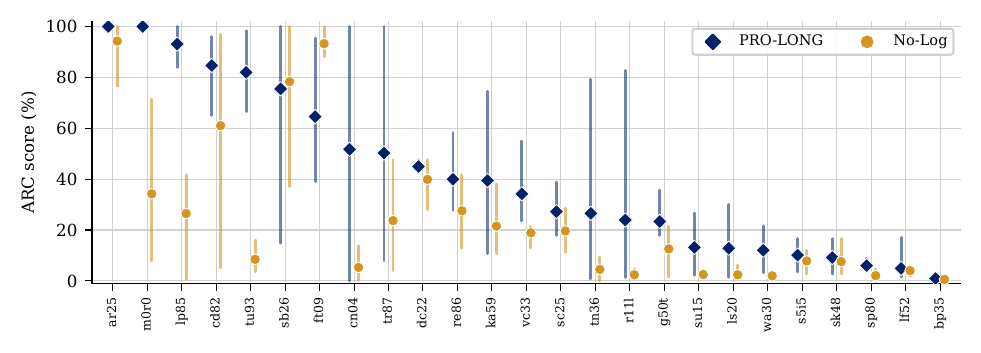}
\caption{\textbf{Score breakdown across the 25 games, PRO-LONG vs.\ \inprompt{}} for GPT 5.5 runs, sorted by PRO-LONG mean. Markers show the pass@1 average per game and bars span the minimum and maximum score over $n=5$ runs. PRO-LONG matches or exceeds \inprompt{} scores on nearly every game.}
\label{fig:pergame_bars}
\end{figure}

\paragraph{Specific game performance for ARC-AGI-3.}
Figure~\ref{fig:pergame_bars} provides the full game-by-game comparison on the ARC-AGI-3 public set; we discuss the games where programmatic memory matters most in \S\ref{sec:general-ablations}.

\paragraph{Ablating no-log format.} Our no-log baseline places each board in the user prompt, from which the agent can save information to workspace files, as described in \S\ref{sec:arch}. We ablate this choice by repeating the accessed-context experiment in Figure~\ref{fig:context_access} with a variant that instead retrieves the current board through a read-only \texttt{observe} MCP tool. Thus, the only intended difference is how the current observation is delivered. The two variants follow the same overall trend. MCP scores 3.0 points higher at the default threshold, but degrades more as the compaction threshold decreases, falling below in-prompt at both 128K and 64K.

\begin{figure}[h]
\centering
\includegraphics[width=0.4\linewidth]{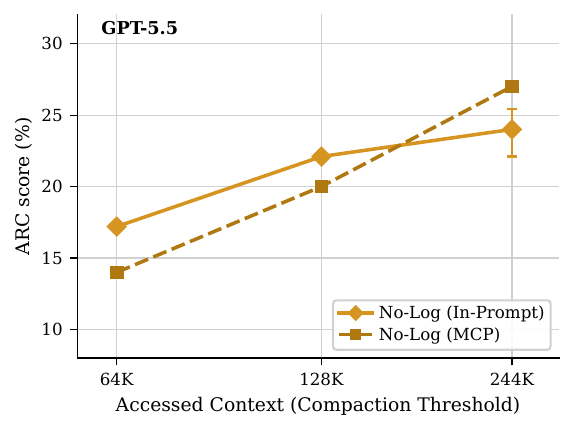}
\caption{\textbf{No-log baseline ablation.} Using the same setup as Figure~\ref{fig:context_access}, we compare the default no-log condition, where the board is placed in the user prompt, with a variant that retrieves it through a read-only \texttt{observe} MCP tool.}
\label{fig:nolog_ablation}
\end{figure}

\begin{figure}[h]
\centering
\includegraphics[width=0.75\linewidth]{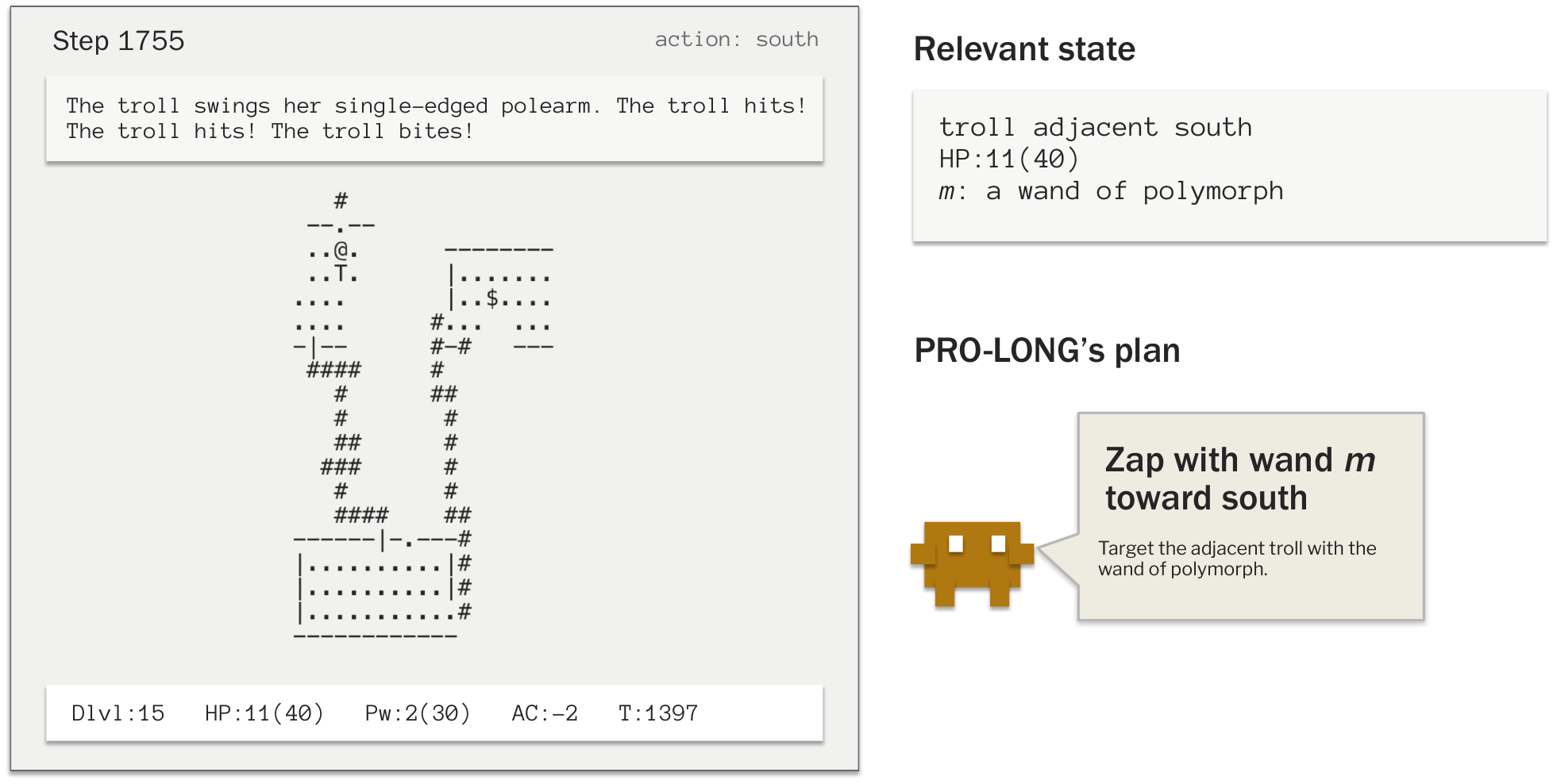}
\caption{\textbf{An example NetHack state from a PRO-LONG episode.} An adjacent troll attacks the agent, leaving it at 11/40 HP (left). The PRO-LONG agent responds by zapping its wand of polymorph south at the troll (right).}
\label{fig:nethack-setup}
\end{figure}

\begin{figure}[h]
\centering
\includegraphics[width=\textwidth]{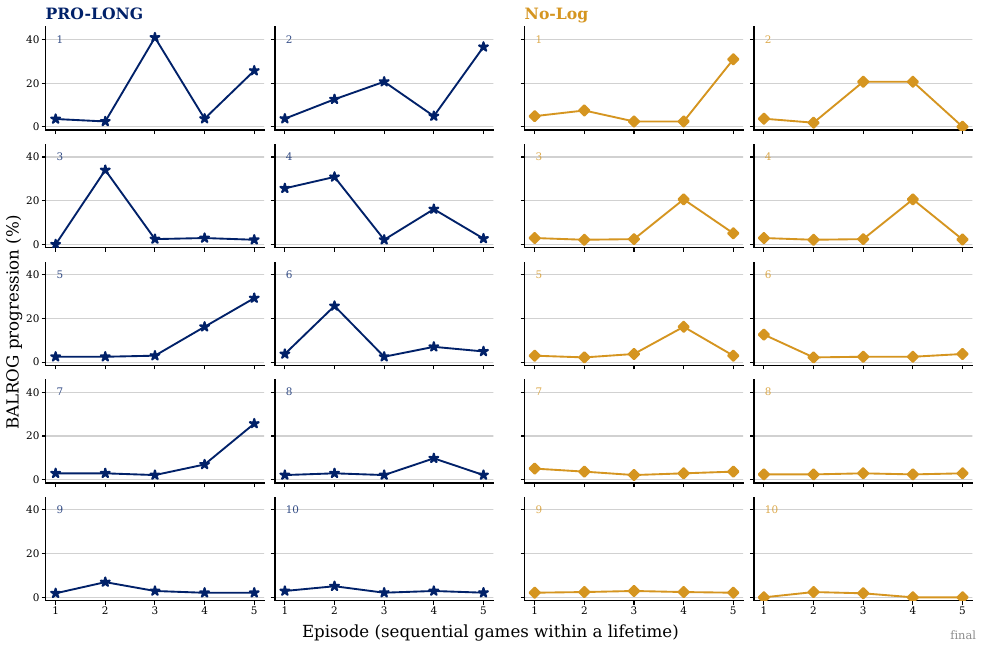}
\caption{\textbf{Every GPT-5.6 + Codex lifetime.} BALROG progression for each of the five sequential episodes in all 10 PRO-LONG (left) and 10 \InPrompt{} (right) lifetimes, ordered by best episode.}
\label{fig:nethack-lifetimes-codex}
\end{figure}

\begin{figure}[h]
\centering
\includegraphics[width=\textwidth]{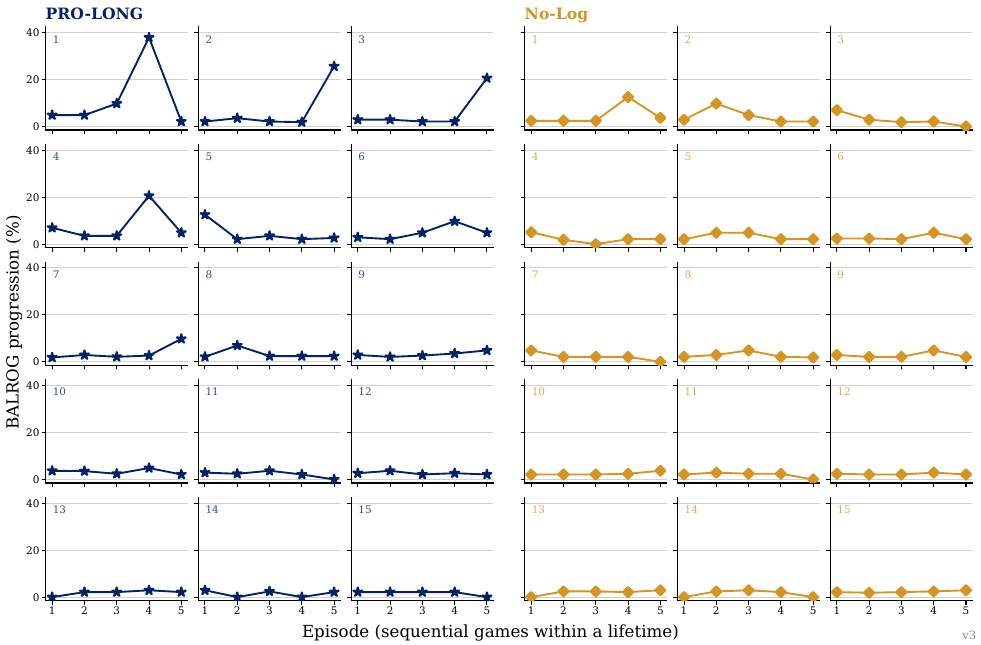}
\caption{\textbf{Every Kimi + Mini-SWE lifetime.} BALROG progression for each of the five sequential episodes in all 15 PRO-LONG (left) and 15 \InPrompt{} (right) lifetimes, ordered by best episode.} %
\label{fig:nethack-lifetimes}
\end{figure}

\paragraph{Individual NetHack lifetimes.} Figures~\ref{fig:nethack-lifetimes-codex} and~\ref{fig:nethack-lifetimes} show every GPT-5.6 + Codex and Kimi + Mini-SWE lifetime behind the means in Figure~\ref{fig:nethack-bestk}. Several PRO-LONG lifetimes score below 5\% in their first one or two episodes and then reach 20--40\%, whereas \InPrompt{} lifetimes mostly stay near the floor.

\section{Full Prompts}
\label{app:prompts}
Values in braces are filled at runtime. Codex takes agent instructions from
\texttt{AGENTS.md}, while Claude Code loads them from \texttt{CLAUDE.md}. Mini-SWE
receives its adapted instructions as a system prompt. We refer to these collectively
as agent instructions below. Each backend also receives the user prompts shown for
its condition.

\subsection{Shared Components}

\subsubsection{Action Space}
\label{app:actions}

\begin{promptbox}[Available Actions]
- ACTION1 --- Up
- ACTION2 --- Down
- ACTION3 --- Left
- ACTION4 --- Right
- ACTION5 --- Spacebar / interact
- ACTION6(x,y) --- Click at column x (0-63), row y (0-63)
- ACTION7 --- Undo
- RESET --- Reset level (actions still count)
\end{promptbox}

\subsubsection{Color Map}

\begin{promptbox}[Color Map (ASCII Mode)]
**Color Map (ASCII character -> color):**
COLOR\_MAP = \{
    '\$': 'White', '8': 'Off-White', '\#': 'Light Gray', 'h': 'Gray',
    'q': 'Off-Black', 'O': 'Black', 'C': 'Magenta', 'z': 'Light Magenta',
    'n': 'Red', 'f': 'Blue', '(': 'Light Blue', 'G': 'Yellow',
    '-': 'Orange', '>': 'Maroon', 'I': 'Green', '"': 'Purple',
\}
\end{promptbox}

\subsection{PRO-LONG}
\label{app:system-fulllog}

\subsubsection{Agent Instructions}

\begin{promptbox}[Agent Instructions (PRO-LONG)]
You are a coding agent playing a grid-based puzzle game by writing Python action plans.

Your primary objective is to solve all levels in the game. Your secondary objective is to minimize total cumulative actions used.

`./logs.txt` is the game log: action headers, tool calls, board states, and your own prior analyses. {log\_window\_desc} Parse it **programmatically**, as reading full 64x64 board states from prompt can introduce precision errors.{cross\_turn\_hint}

**Tools**: Read, Write, Edit, Bash, Grep, Glob.

**Workspace**: `./` persists across calls. `actions.json` is cleared each call; other files accumulate. Feel free to save notes, state, or helper functions.

**Log markers**:
    [INITIAL BOARD STATE] --- the grid at the start (after Action 0 header)
    [POST-ACTION BOARD STATE] --- the grid after each action (when the log includes action history)
    [frame 1/N] ... [settled] --- animation frames if the log includes them; the grid following [settled] (or the only grid, if there are no [frame] markers) is the committed state

**Game structure and strategy**:
- Score increase means that a level was cleared.
- Most games have a step budget or timer mechanism, which will cause a level reset if exceeded.
- For parsing the boards, programmatic options include identifying connected components (color, position, size, shape) and forming testable hypotheses about what each represents (player, walls, goals, UI, etc.).

**Response format**: a strategic briefing, then
[PLAN]
<2-3 sentence action plan>

**Write `./actions.json`** with a JSON object \{"actions": ["ACTION6(30,40)", "ACTION1", "RESET"]\} --- a list of 1--{action\_cap} actions to execute in order. Entries beyond {action\_cap} are discarded. Prefer short lists (1--2 actions) when testing a new hypothesis so you see the result before committing further; scale up toward {action\_cap} for proven sequences.

**Actions available in this game**:
{actions\_section}

The runner executes the list in order, then calls you again with the updated log.
\end{promptbox}

\subsubsection{User Prompts}

\begin{promptbox}[Initial User Prompt]
Read the full game log at ./logs.txt

This is the first analysis. Analyze the board state and write ./actions.json with your first set of actions.
\end{promptbox}

\begin{promptbox}[Resume User Prompt]
Read the full game log at ./logs.txt

Recent actions and boards are at the end of the log; what changed since the last call (new moves, score transitions, plan adherence) can be informative. Check ./ for anything you saved previously, then write a new ./actions.json.
\end{promptbox}

\subsubsection{Mini-SWE Setup}

For the Kimi experiments, Mini-SWE receives the PRO-LONG instructions above as its
system prompt. Mini-SWE exposes the Bash tool, replaces the tool line below, omits the response format block, and appends the command instructions below. The initial and resume user prompts are unchanged.

\begin{promptbox}[Mini-SWE Prompt Addendum]
**Tool**: Bash. Use shell commands to read, write, edit, and search files.

\#\# Command execution rules

Use the Bash tool to inspect and edit the workspace. Each command starts in a fresh shell.

When the action batch is ready, write the final analysis and optional [PLAN] to ./last\_message.txt. Then submit with this command alone:

echo COMPLETE\_TASK\_AND\_SUBMIT\_FINAL\_OUTPUT
\end{promptbox}

\subsection{\InPrompt{}}
\label{app:system-inprompt}

\subsubsection{Agent Instructions}

\begin{promptbox}[Agent Instructions (\InPrompt{})]
You are a coding agent playing a grid-based puzzle game by writing Python action plans.

Your primary objective is to solve all levels in the game. Your secondary objective is to minimize total cumulative actions used.

The current 64x64 board is included in the user prompt each turn. The workspace does not contain a game log, but files you create there persist across calls.

**Tools**: Read, Write, Edit, Bash, Grep, Glob.

**Workspace**: `./` persists across calls. `actions.json` is cleared each call; other files accumulate. Feel free to save notes, state, or helper functions.

**Prompt markers** (on the user turn):
    [CURRENT BOARD STATE] --- the live 64x64 grid

**Game structure and strategy**:
- Score increase means that a level was cleared.
- Most games have a step budget or timer mechanism, which will cause a level reset if exceeded.
- For parsing the boards, programmatic options include identifying connected components (color, position, size, shape) and forming testable hypotheses about what each represents (player, walls, goals, UI, etc.).

**Response format**: a strategic briefing, then
[PLAN]
<2-3 sentence action plan>

**Write `./actions.json`** with a JSON object \{"actions": ["ACTION6(30,40)", "ACTION1", "RESET"]\} --- a list of 1--{action\_cap} actions to execute in order. Entries beyond {action\_cap} are discarded. Prefer short lists (1--2 actions) when testing a new hypothesis so you see the result before committing further; scale up toward {action\_cap} for proven sequences.

**Actions available in this game**:
{actions\_section}

The runner executes the list in order, then calls you again with an updated board.
\end{promptbox}

\subsubsection{User Prompts}
\label{app:user-prompts}

\begin{promptbox}[Initial User Prompt]
[CURRENT BOARD STATE]
{board}

This is the first analysis. Analyze the board above and write ./actions.json with your first set of actions.
\end{promptbox}

\begin{promptbox}[Resume User Prompt]
[CURRENT BOARD STATE]
Score: {score} | Action: {action\_num} | Level: {level}
Last actions: {last\_actions}

{board}

What ran since the last call is listed above; check ./ for anything you saved previously, then write a new ./actions.json.
\end{promptbox}

\subsection{MCP}
\label{app:system-readonly-mcp}

\subsubsection{Agent Instructions}

\begin{promptbox}[Agent Instructions (MCP)]
You are a coding agent playing a grid-based puzzle game by writing Python action plans.

Your primary objective is to solve all levels in the game. Your secondary objective is to minimize total cumulative actions used.

The game observation is available through the read-only `observe` MCP tool. The workspace does not contain a game log, but files you create there persist across calls.

**Tools**: Read, Write, Edit, Bash, Grep, Glob.

**Workspace**: `./` persists across calls. `actions.json` is cleared each call; other files accumulate. Feel free to save notes, state, or helper functions.

**Observation tool**:
    `observe` --- returns the current settled board and, when available, the executed actions and animation frames from the preceding batch

**Game structure and strategy**:
- Score increase means that a level was cleared.
- Most games have a step budget or timer mechanism, which will cause a level reset if exceeded.
- For parsing the boards, programmatic options include identifying connected components (color, position, size, shape) and forming testable hypotheses about what each represents (player, walls, goals, UI, etc.).

**Response format**: a strategic briefing, then
[PLAN]
<2-3 sentence action plan>

**Write `./actions.json`** with a JSON object \{"actions": ["ACTION6(30,40)", "ACTION1", "RESET"]\} --- a list of 1--{action\_cap} actions to execute in order. Entries beyond {action\_cap} are discarded. Prefer short lists (1--2 actions) when testing a new hypothesis so you see the result before committing further; scale up toward {action\_cap} for proven sequences.

**Actions available in this game**:
{actions\_section}

The runner executes the list in order, then calls you again with an updated board.
\end{promptbox}

\subsubsection{User Prompts}

\begin{promptbox}[Initial User Prompt]
Call `observe` to inspect the game.

This is the first analysis. Analyze the board returned by the tool and write ./actions.json with your first set of actions.
\end{promptbox}

\begin{promptbox}[Resume User Prompt]
Score: {score} | Action: {action\_num} | Level: {level}
Last actions: {last\_actions}

Call `observe` to inspect the current boards. Check ./ for anything you saved previously, then write a new ./actions.json.
\end{promptbox}

\subsubsection{MCP Interface}

\begin{promptbox}[MCP Server Instruction]
Inspect the game observation state.
\end{promptbox}

\begin{promptbox}[`Observe` Tool]
Name: observe

Arguments: none

Description: Return the current board and preceding action frames.

The first call returns:

\{
  "board": "<current settled board>"
\}

Later calls return:

\{
  "board": "<current settled board>",
  "executed\_actions": ["ACTION1", "ACTION2"],
  "frames": [
    \{
      "action\_index": 1,
      "frame\_index": 1,
      "settled": false,
      "board": "<frame>"
    \}
  ]
\}
\end{promptbox}

\subsection{NetHack Prompts}
\label{app:nethack-prompts}
The NetHack coding agent reads the following instructions from \texttt{AGENTS.md}. The
environment provides its action list and rules separately in \texttt{instructions.txt}.

\begin{promptbox}[Shared Agent Instructions]
\setlength{\parskip}{0pt}
\{state\_access\}
\{workspace\_access\}

Status: Dlvl is dungeon depth, \$ is gold, HP is current/maximum health, Pw is spell
power, AC is armor class (lower is better), Xp is experience level/points, and T is game turn.

NetHack reminders:
- NetHack is permadeath: reaching 0 HP ends the run. Never take the up staircase on
  Dlvl 1; it exits the dungeon and also ends the run.\par
- Monsters can act after each turn. They may pursue, attack at range, inflict conditions,
  or act faster than you. Avoid trading attacks at low HP when escape or recovery is available.\par
- Do not rest or search repeatedly while a hostile is nearby.\par
- A pickaxe is primarily a digging tool. Digging may take several turns while nearby
  monsters continue acting, so avoid digging near hostiles and re-wield your combat
  weapon afterward.\par
- Hunger progresses from Hungry to Weak to Fainting; eat before Weak. Corpses can rot,
  so only eat newly killed or known-safe corpses. Burdened characters move more slowly.\par
- Prayer can resolve emergencies but has a long cooldown. Engraving Elbereth can deter
  many melee attackers.\par
- Shop merchandise becomes unpaid when picked up. Pay for it or return it before leaving,
  or the shopkeeper may become hostile.\par
- The Gnomish Mines and Sokoban are side branches, so progress can require returning to
  the main dungeon. Cave levels with gnomes and dwarves indicate the Mines; a town of
  shops and a temple is Minetown; boulder puzzles indicate Sokoban.\par

When ready, import submit from submit.py and call it once with the chosen actions.
Submit at most 30 actions. They execute in order without intermediate model feedback.
Use 1-2 actions when testing something uncertain; scale toward 30 for a safe, proven sequence.
The runner stops the batch and asks again if you take damage or change dungeon level.
It also stops if a new monster appears or one action advances multiple game turns.
\end{promptbox}

\begin{promptbox}[PRO-LONG: State and Workspace Access]
Read instructions.txt for the environment rules and valid actions. logs.txt contains the
complete game history, including the current observation at the end. Read and search it
rather than relying on memory.

Use Python and shell tools to analyze the game state and available history programmatically
when useful. You may save notes, state, or helper functions in the workspace across calls.
\end{promptbox}

\begin{promptbox}[No-Log: State and Workspace Access]
Read instructions.txt for the environment rules and valid actions. The current observation
is included in each user message.

You may save notes and helper functions in the workspace across calls.
\end{promptbox}

For multi-attempt runs, the next user prompt also includes:
\begin{promptbox}[Attempt Restart]
A new attempt has started from the same seed. Your goal remains to maximize your chance of
eventually ascending. Use what you learned from previous attempts to avoid repeating
mistakes and improve your strategy.
\end{promptbox}

\begin{promptbox}[PRO-LONG: User Prompt]
The game state has been appended to logs.txt. Use Python or shell tools when useful to
analyze the current state or relevant history, for example by parsing board coordinates,
tracking explored routes and stairs, comparing status over time, or solving spatial puzzles
(e.g. via BFS). When ready, submit the next action batch.
\end{promptbox}

\begin{promptbox}[No-Log: User Prompt]
[CURRENT OBSERVATION]
\{observation\}

Inspect the current game state and submit the next action batch.
\end{promptbox}

\begin{promptbox}[NetHack Environment Instructions]
\setlength{\parskip}{0pt}
You are an agent playing NetHack. The following are the common named actions you can take in the game, followed by a short description of each action:

\{action\_strings\}.

Tips:
- When the message asks for a completion, such as: "What do you want to eat? [d or ?*]", you should respond with a single character corresponding to the item you want to eat/use.\par
    - For example, "What do you want to eat? [dgh or ?*]" -> Possible answers are "d", "g", or "h" to eat the associated food.\par
- NetHack reads one key per action. Enter counts one digit at a time. To type free text, submit one action per character followed by "enter". For example, engraving Elbereth is ["engrave", "minus", "E", "l", "b", "e", "r", "e", "t", "h", "enter"].\par
- When the message asks for a direction, such as: "In what direction?" you should respond with a direction.\par
- When the message has --More-- at the end, your next action should be "more" to see the rest of the message.\par
- Explore the environment to find the stairs down to the next level.\par
- Always carefully read the last message to understand the current state of the game and decide your next action accordingly.\par
- If you keep moving in the same direction, you will eventually hit a wall and stop moving. Your message might be: "It's solid stone", or "It's a wall". Change your action to move in another direction to continue exploring the environment.\par
- Read the language observation carefully and look at ascii map or image observation provided to decide the next action to take and where to move next.\par
- You can attack monsters by moving into them.\par

Your goal is to maximize your chance of eventually ascending. Make sustained progress by
surviving, gaining experience and useful equipment, and descending when prepared. Depth
matters, but so do survival, experience, and equipment.
\end{promptbox}

\end{document}